%% file: main.tex
\pdfoutput=1

\documentclass[11pt]{article}

\usepackage{naacl2021}

\usepackage{times}
\usepackage{latexsym}

\usepackage[T1]{fontenc}

\usepackage[utf8]{inputenc}

\usepackage{microtype}

\usepackage{float}

\usepackage{booktabs}
\usepackage{xspace}
\usepackage{graphicx}
\usepackage{amsmath, amsthm, amssymb}

\newcommand{\bm}{\textsc{BM25}\xspace}
\newcommand{\bert}{\textsc{BERT}\xspace}
\newcommand{\tapas}{\textsc{TAPAS}\xspace}
\newcommand{\dtr}{\textsc{DTR}\xspace}

\newcommand{\nqt}{\textsc{NQ-Tables}\xspace}

\newcommand{\err}[1]{ $\pm$ #1}

\title{Open Domain Question Answering over Tables via Dense Retrieval}

\author{Jonathan Herzig\thanks{\; Work completed while interning at Google.}$\mathbf{^{*,1}}$, Thomas M{\"u}ller\textsuperscript{2}, Syrine Krichene\textsuperscript{2}, Julian Martin Eisenschlos\textsuperscript{2} \\
\\ $^1$School of Computer Science, Tel-Aviv University\\
\texttt{\large jonathan.herzig@cs.tau.ac.il}\\\\
$^2$Google Research \\
\texttt{\large \{thomasmueller,syrinekrichene,eisenjulian\}@google.com} \\
}

\date{}

\begin{document}
\maketitle

\input 0_abstract
\input 1_introduction
\input 2_setup
\input 3_dense_table_retrieval
\input 4_qa

\input 5_dataset
\input 6_experiments
\input 8_conclusion
\input 10_acknowledgments

\bibliography{references}
\bibliographystyle{acl_natbib}

\clearpage
\input A_appendix

\end{document}

%% file: 0_abstract.tex
\begin{abstract}

Recent advances in open-domain QA have led to strong models based on dense retrieval, but only focused on retrieving textual passages.
In this work, we tackle open-domain QA over tables for the first time, and show that retrieval can be improved by a retriever designed to handle tabular context.
We present an effective pre-training procedure for our retriever and improve retrieval quality with mined hard negatives.
As relevant datasets are missing, we extract a subset of \textsc{Natural Questions} \cite{47761} into a Table QA dataset. We find that our retriever improves retrieval results from $72.0$ to $81.1$ recall@10 and end-to-end QA results from $33.8$ to $37.7$ exact match, over a BERT based retriever.

\end{abstract}

%% file: 1_introduction.tex
\section{Introduction}

Models for question answering (QA) over tables usually assume that the relevant table is given during test time. This applies for semantic parsing (e.g., for models trained on \textsc{SPIDER} \cite{yu-etal-2018-spider})
and for end-to-end QA \cite{neelakantan2016neural,herzig-etal-2020-tapas}. While this assumption simplifies the QA model, it is not realistic for many use-cases where the question is asked through some open-domain natural language interface, such as web search or a virtual assistant.

In these open-domain settings, the user has some information need, and the corresponding answer resides in some table in a large corpus of tables. The QA model then needs to utilize the corpus as an information source, efficiently search for the relevant table within, parse it, and extract the answer. 

Recently, much work has explored open-domain QA over a corpus of textual passages \cite[\emph{inter alia}]{chen-etal-2017-reading,sun-etal-2018-open, yang-etal-2019-end,lee-etal-2019-latent}. These approaches usually follow a two-stage framework: (1) a retriever first selects a small subset of candidate passages relevant to the question, and then (2) a machine reader examines the retrieved passages and selects the correct answer.
While these approaches work well on free text, it is not clear whether they can be directly applied to tables, as tables are semi-structured, and thus different than free text.

\begin{figure}[t]
\centering
\includegraphics[width=1.0\columnwidth]{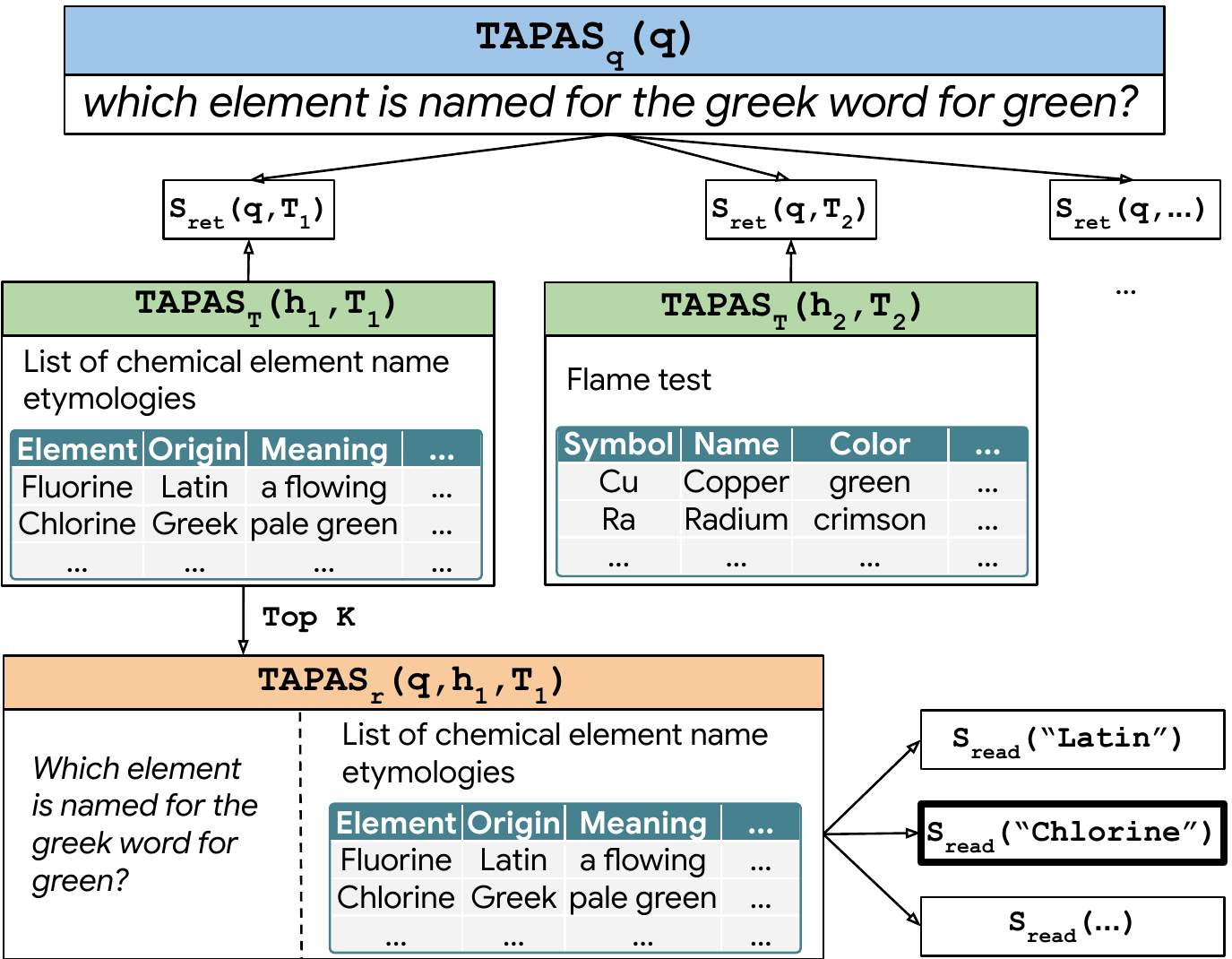}
\caption{An overview of our approach. A dense table retriever scores the question against all tables and outputs the top $K$ tables ($K=1$ in this example), and a reader selects the answer out of the top $K$ tables.}
\label{fig:full_model}
\end{figure}

In this paper we describe the first study to tackle open-domain QA over tables, and focus on modifying the retriever. We follow the two-step approach of a retriever model that retrieves a small set of candidate tables from a corpus, followed by a QA model (Figure 
\ref{fig:full_model}).
Specifically, we utilize dense retrieval approaches targeted for retrieving passages \cite{lee-etal-2019-latent,guu2020realm,karpukhin2020dense}, and modify the retriever to better handle tabular contexts. 
We present a simple and effective pre-training procedure for our retriever, and further improve its performance by mining hard negatives using the retriever model. Finally, as relevant open domain datasets are missing, we process \textsc{Natural Questions}~\cite{47761} and extract 11K examples where the answer resides in some table.
Our model and data generation code as well as the pre-trained model are publicly available at \url{https://github.com/google-research/tapas}.

%% file: 2_setup.tex
\section{Setup}

We formally define open domain extractive QA over tables as follows. We are given a training set of $N$
examples $\mathcal{D}_{\text{train}}= \{(q_i,T_i,a_i)\}_{i=1}^{N}$, where $q_i$ is a question, $T_i$ is a table where the answer $a_i$ resides, and a corpus of $M$ tables $\mathcal{C}=\{T_i\}_{i=1}^{M}$. 
The answer $a_i$ is comprised of one or more spans of tokens in $T_i$. Our goal is to learn a model that given a new question $q$ and the corpus $\mathcal{C}$ returns the correct answer $a$.

Our task shares similarities with open domain QA over documents \cite{chen-etal-2017-reading, yang-etal-2019-end,lee-etal-2019-latent}, where the corpus $\mathcal{C}$ consists of textual passages extracted from documents instead of tables, and the answer is a span that appears in some passage in the corpus.
As in these works, dealing with a large corpus (of tables in our setting), requires relevant context retrieval. Naively applying a QA model, for example \tapas \cite{herzig-etal-2020-tapas}, over each table in the large corpus is not practical because inference is too expensive.

To this end we break our system into two independent steps. First, an efficient table retriever component selects a small
set of candidate tables $\mathcal{C}_R$ from a large corpus of tables $\mathcal{C}$. Second, we apply a QA model to extract the answer $a$ given the question $q$ and the candidate tables $\mathcal{C}_R$.

%% file: 3_dense_table_retrieval.tex
\section{Dense Table Retrieval}
\label{sec:dense_ret}

In this section we describe our dense table retriever (\dtr), which retrieves a small set of $K$ candidate tables $\mathcal{C}_R$ given a question $q$ and a corpus $\mathcal{C}$. In this work we set $K=10$ and take $\mathcal{C}$ to be the set of all tables in the dataset we experiment with (see \S\ref{sec:experiments}).

As in recent work for open domain QA on passages \cite{lee-etal-2019-latent,guu2020realm,karpukhin2020dense,chen2020open,oguz2020unified}, we also follow a dense retrieval architecture. As tables that contain the answer to $q$  do not necessarily include tokens from $q$, a dense encoding can better capture similarities between table contents and a question.

For training \dtr, we leverage both in-domain training data $\mathcal{D}_{\text{train}}$, and automatically constructed pre-training data $\mathcal{D}_{\text{pt}}$ of text-table pairs (see below). 

\paragraph{Retrieval Model} In this work we focus on learning a retriever that can represent tables in a meaningful way, by capturing their specific structure. Traditional information retrieval methods such as \bm are targeted to capture token overlaps between a query and a textual document, and other dense encoders are pre-trained language models (such as \bert) targeted for text representations.

Recently, \newcite{herzig-etal-2020-tapas} proposed \tapas, an encoder based on \bert, designed to contextually represent text and a table jointly. \tapas includes table specific embeddings that capture its structure, such as row and column ids. 
In \dtr, we use \tapas to represent both the query $q$ and the table $T$. For efficient retrieval during inference we use two different \tapas instances (for $q$ and for $T$), and learn a similarity metric between them as \newcite{lee-etal-2019-latent,karpukhin2020dense}.

More concretely,
the \tapas encoder $\texttt{TAPAS}(x_1, [x_2])$ takes one or two inputs as arguments, where $x_1$ is a string and $x_2$ is a flattened table.
We then define the retrieval score as the inner product of dense vector representations of the question $q$ and the table $T$:
\begin{align*}
    &h_q = \mathbf{W_q}\texttt{TAPAS}_{\text{q}}(q)\texttt{[CLS]} \\
    &h_T = \mathbf{W_T}\texttt{TAPAS}_{\text{T}}(\texttt{title}(T), T)\texttt{[CLS]} \\
    &\texttt{S}_\text{ret}(q, T) = h^T_q h_T,
\end{align*}
where $\texttt{TAPAS}(\cdot)\texttt{[CLS]}$ returns the hidden state for the \texttt{CLS} token, $\mathbf{W_q}$ and $\mathbf{W_T}$ are matrices that project the
\tapas output into $d=256$ dimensional vectors, and $\texttt{title}(T)$ is the page title for table $T$. We found the table's page title to assist in retrieving relevant tables, which is also useful for Wikipedia passage retrieval \cite{lee-etal-2019-latent}.

\paragraph{Training}
The goal of the retriever is to create a vector space such that relevant pairs of questions and tables will have smaller distance (which results in a large dot product) than the irrelevant pairs, by learning an
embedding. To increase the likelihood of gold $(q, T)$ pairs, we train the retriever with in-batch negatives \cite{gillick-etal-2019-learning,Henderson2017EfficientNL,karpukhin2020dense}. Let $\{(q_i,T_i)\}_{i=1}^{B}$ be a batch of $B$ examples from $\mathcal{D}_{\text{train}}$, where for each $q_i$, $T_i$ is the gold table to retrieve, and for each $j\neq i$ we treat $T_j$ as a negative. We now define the likelihood of the gold table $T_i$ as:    
\begin{align*}
    p(T_i | q_i)=\frac{\exp[\texttt{S}_\text{ret}(q_i,T_i)]}{\sum_{j=1}^B \exp[\texttt{S}_\text{ret}(q_i,T_j)]}.
\end{align*}

To train the model efficiently, we define $\mathbf{Q}$ and $\mathbf{T}$ to be a $B\times d$ matrix that hold the representations for questions and tables respectively. Then, $\mathbf{S}=\mathbf{Q}\mathbf{T}^T$ gives an $B\times B$ matrix where the logits from the gold table are on the diagonal. We then train using a row-wise cross entropy loss where the labels are a $B\times B$ identity matrix. 

\paragraph{Pre-training}
One could train our retriever from scratch, solely relying on a sufficiently large in-domain training dataset $\mathcal{D}_{\text{train}}$. 
However, we find performance to improve after using a simple pre-training method for our retriever. \newcite{lee-etal-2019-latent} suggest to pre-train a textual dense retriever using an Inverse Cloze Task (ICT). In ICT, the goal is to predict a context given a sentence $s$. The context is a passage that originally contains $s$, but with $s$ masked. The motivation is that the relevant context should be semantically similar to $s$, and should contain information missing from $s$.

Similarly, we posit that a table $T$ that appears in close proximity to some text span $s$ is more relevant to $s$ than a random table. To construct a set $\mathcal{D}_{\text{pt}}= \{(s_i,T_i)\}_{i=1}^{M}$ that consists of $M$ pre-training pairs $(s,T)$, we use the pre-training data from \newcite{herzig-etal-2020-tapas}.
They extracted text-table pairs from 6.2M Wikipedia tables, where text spans were sampled from the table caption, page title, page description, segment title and text of the segment the table occurs in. This resulted in a total of 21.3M text-table $(s,T)$ pairs. While \newcite{herzig-etal-2020-tapas} uses extracted $(s,T)$ pairs for pre-training \tapas with a masked language modeling objective, we pre-train \dtr from these pairs, with the same objective used for in-domain data. 

\paragraph{Hard Negatives}
Following similar work \cite{gillick-etal-2019-learning,karpukhin2020dense,xiong2020approximate},
we use an initial retrieval model to extract the most similar tables from $\mathcal{C}$ for each question in the training set.
From this list we discard each table that does contain the reference answer to remove false negatives.
We use the highest scoring remaining table as a particular hard negative.

Given the new triplets of question, reference table and mined negative table, we train a new model using a modified version of the in-batch negative training discussed above.
Given $\mathbf{Q}$ and $\mathbf{S}$ as defined above and a new matrix $\mathbf{N}$ ($B\times d$)
that holds the representations of the negative tables, $\mathbf{S'}=\mathbf{Q}\mathbf{N}^T$ gives another $B\times B$ matrix that we want to be small in value (possibly negative). If we concatenate $\mathbf{S}$ and $\mathbf{S'}$ row-wise
we get a new matrix for which we can perform the same cross entropy training as before.
The label matrix is now obtained by concatenating an identity matrix row-wise with a zero matrix.

\paragraph{Inference}
During inference time, we apply the
table encoder $\texttt{TAPAS}_{\text{T}}$ to all the tables $T \in \mathcal{C}$ offline.
Given a test question $q$, we derive its representation $h_q$ and retrieve the top $K$ tables with representations closest to $h_q$. 

In our experiments, we use exhaustive search to find the top $K$ tables, but to scale to large corpora, fast maximum inner product search using existing tools such as \textsc{FAISS} \cite{JDH17} and \textsc{ScaNN} \cite{scann} could be used, instead.

%% file: 4_qa.tex
\section{Question Answering over Tables}

A reader model is used to extract the answer $a$ given the question $q$ and $K$ candidate tables. 
The model scores each candidate and at the same time extracts a suitable answer span from the table.
Each table and question are jointly encoded using a TAPAS model.
The score is a simple logistic loss based on the \texttt{CLS} token, as in ~\citet{eisenschlos-etal-2020-understanding}.

The answer span extraction is modeled as a soft-max over all possible spans up to a certain length.
Spans that are located outside of a table cell or that cross a cell are masked.
Following \newcite{46490,lee-etal-2019-latent}, the span representation is the concatenation of the contextual representation of the first and
last token in the span $s$: 
\begin{align*}
    &h_{start} = \texttt{TAPAS}_r(q, \texttt{title}(T), T)[\texttt{START}(s)] \\
    &h_{end} = \texttt{TAPAS}_r(q, \texttt{title}(T), T)[\texttt{END}(s)] \\
    &\texttt{S}_\text{read}(q, T) = \texttt{MLP}([h_{start}, h_{end}]).
\end{align*}
The training and test data are created by running a retrieval model.
We extract the $K=10$ highest scoring candidate tables for each question.
At training time we add the reference table if it is missing from the candidates.

At inference time all table candidates are processed and the answer of the candidate with the highest score is returned as the predicted answer.

%% file: 5_dataset.tex
\section{Dataset}

We create a new English dataset called \textbf{\nqt} from \textsc{Natural Questions} \cite{47761} (NQ).
Concurrently with this work, \citet{zayats2021representations} study a similar subset of NQ
but without the retrieval aspect.

NQ was designed for question answering over Wikipedia articles.
The $320K$ questions are mined from real Google search queries and the answers are spans in Wikipedia articles identified by annotators.
Although the answers for most questions appear in textual passages, we identified 12K examples where the answer resides in a table, and can be used as a QA over tables example.
To this end, we form \nqt that consists of $(q,T,a)$ triplets from these examples.
Tables are extracted from the article's HTML, and are normalized by transposing \emph{infobox} tables.

We randomly split the original NQ train set into train and dev (based on a hash of the page title) and use all questions from the original NQ dev set as our test set.
To construct the corpus $\mathcal{C}$, we extract all tables that appear in articles in all NQ sets.

NQ can contain the same Wikipedia page in different versions which leads to many almost identical tables.
We merge close duplicates using the following procedure.
For all tables that occur on the same Wikipedia page we flatten the entire table content, tokenize it and compute $l_2$ normalized uni-gram vectors of the token counts of each table.
We then compute the pair-wise cosine similarity of all tables.
We iterate over the table pairs in decreasing order of similarity and attempt to merge them into clusters.
This is essentially a version of single link clustering.
In particular, we will merge two tables if 
the similarity is $> 0.91$, they do not occur on the same version of the page,
their difference is rows is at most 2 and they have the same number of columns.

Dataset sizes are given in the following table: 

\begin{center}
\resizebox{0.6\columnwidth}{!}{
\begin{tabular}{lcccc}
train & dev   & test & corpus $\mathcal{C}$ \\
\toprule
9,594 & 1,068 & 966  & 169,898 \\
\end{tabular}}
\end{center}

%% file: 6_experiments.tex
\section{Experiments}
\label{sec:experiments}

\begin{table}[t]
\begin{center}
\resizebox{0.8\columnwidth}{!}{
\begin{tabular}{lccc}
Model &    R@1 & R@10 & R@50 \\
\toprule
BM25       & 16.77 & 40.06 & 58.39 \\
DTR-Text   & 32.90 & 72.00 & 86.86 \\
DTR-Schema & 34.36 & 74.24 & 88.37 \\
\midrule
DTR         & 36.24 & 76.02 & 90.25 \\
DTR +hnbm25 & 42.17 & 80.51 & 92.31 \\
DTR +hn     & \textbf{42.42} & \textbf{81.13} & \textbf{92.56} \\
\midrule
DTR -pt &  16.64 & 47.80 & 68.68 \\
\bottomrule
\end{tabular}
}
\end{center}
\caption{Table retrieval results on NQ-TABLES test set. hn: hard negatives from DTR, hnbm25: hard negatives from BM25 baseline, pt: pre-training.
DTR numbers are means over 5 random runs.
}
\label{tab:test_retrieval}
\end{table}

Details about the experimental setup are given Appendix~\ref{app:exp_setup}.

\paragraph{Retrieval Baselines}
We consider the following baselines as alternatives to \dtr.
We use the \textbf{BM25} \cite{robertson2009probabilistic} implementation of Gensim \cite{rehurek_lrec}\footnote{We find that recall improves if the document title and table header tokens are counted multiple times. In all experiments we use a count of 15.}.
To measure if a table-specific encoder is necessary, we implement \textbf{\textsc{DTR-text}}, where the retriever is initialized from \bert \cite{devlin2018BERT} instead of \tapas.
To test whether the content of the table is relevant, we experiment with \textbf{\textsc{DTR-schema}}, where only the headers and title are used to represent tables. 

\paragraph{Retrieval Results}
Table \ref{tab:test_retrieval} shows the test results for table retrieval (dev results are in Appendix~\ref{app:results}). 
We report recall at $K$ (R@K) metrics as the fraction of questions for which the highest scoring $K$ tables contain the reference table.

We find that all dense models that have been pre-trained out-peform the BM25 baseline by a large margin.
The model that uses the \tapas table embeddings (DTR) out-performs the dense baselines by more than 1 point in R@10.
The addition of mined negatives (DTR +hn) yields an additional improvement of more than 5 points.
Mining negatives from DTR works better than mining negatives from BM25 (DTR +hnbm25, +0.6 R@10).

\begin{table}[t]
\begin{center}
\resizebox{\columnwidth}{!}{
\begin{tabular}{llcccc}
Retriever & Reader & EM & F1 & Oracle EM & Oracle F1 \\
\toprule
BM25  & TAPAS  & 21.46 & 28.24 & 29.51 & 40.79 \\
DTR-Text  & BERT & 29.58 & 37.38 & 39.39 & 51.48 \\
DTR-Text  & TAPAS & 33.78 & 43.49 & 42.83 & 56.46 \\
DTR-Schema  & TAPAS & 32.75 & 42.19 & 42.63 & 55.05 \\
\midrule
DTR  & TAPAS  & 35.50 & 45.44 & 46.09 & 59.01 \\
DTR +hnbm25 & TAPAS & 36.61 & 46.74 & 47.46 & 60.72 \\
DTR +hn & TAPAS & \textbf{37.69} & \textbf{47.70} & \textbf{48.20} & \textbf{61.50} \\
\bottomrule
\end{tabular}
}
\end{center}
\caption{QA results on NQ-TABLES test set. 
Numbers are means over 5 random runs.
}
\label{tab:test_qa}
\end{table}

\paragraph{End-to-End QA}
Results for end-to-end QA experiments are shown in Table \ref{tab:test_qa} (dev results are in Appendix~\ref{app:results}).
We use the exact match (EM) and token F1 metrics as implemented in SQUAD \cite{rajpurkar-etal-2016-squad}.\footnote{\url{https://worksheets.codalab.org/rest/bundles/0x6b567e1cf2e041ec80d7098f031c5c9e/contents/blob/}}
We additionally report oracle metrics which are computed on the best answer returned for any of the candidates.

We again find that all dense models out-perform the BM25 baseline.
A TAPAS-based reader out-performs a BERT reader by more than 3 points in EM.
The simple DTR model out-performs the baselines by more than 1 point in EM. Hard negatives from BM25 (+hnbm25) improve DTR's performance by 1 point, while hard negatives from DTR (+hn) improve performance by 2 points. We additionally perform a McNemar's significance test for our proposed model, DTR+hn, and find that it performs significantly better (p<0.05) than all baselines.

\paragraph{Analysis}

Analyzing the best model in Table \ref{tab:test_qa} (DTR +hn) on the dev set,
we find that 29\% of the questions are answered correctly, 14\% require a list answer (which is out of scope for this paper),
12\% do not have any table candidate that contains the answer, for 11\% the model does not select a table that contains the answer, and for
34\% the reader fails to extract the correct span.

We further analyzed the last category by manually annotating 100 random examples.
We find that for 23 examples the answer is partially correct (usually caused by inconsistent span annotations in NQ).
For 11 examples the answer is ambiguous (e.g., the release date of a movie released in different regions).
For 22 examples the table is missing context or does only contain the answer accidentally.
Finally, 44 examples are wrong, usually because they require some kind of table reasoning, like computing the maximum over a column, or using common sense knowledge.

%% file: 8_conclusion.tex
\section{Conclusion}

In this paper
we demonstrated that a retriever designed to handle tabular context can outperform other textual retrievers for open-domain QA on tables.
We additionally showed that our retriever can be effectively pre-trained and improved by hard negatives.
In future work we aim to tackle multi-modal open-domain QA, combining passages and tables as context.

%% file: 10_acknowledgments.tex
\section*{Acknowledgments}

We would like to thank William Cohen, Sewon Min, Yasemin Altun and the anonymous reviewers for their constructive feedback, useful comments and suggestions. This work was completed in partial fulfillment for the PhD degree of the first author, which was also supported by a Google PhD fellowship.

%% file: A_appendix.tex
\appendix

\section{Experimental Setup}
\label{app:exp_setup}

The \dtr model uses a batch size of 256. 
We pre-trained the question and table encoders for 1M steps, and fine-tuned them for a maximum of 200,000 steps, with a learning rate of 1.25e-5 using Adam and linear scheduling with warm-up and dropout rate 0.2. The hyper-parameter values were selected based on the values used by \newcite{herzig-etal-2020-tapas} on the SQA dataset.
We evaluate \dtr performance using recall@k, and do early stopping according to recall@10 on the dev set.
We use only the tables that appear in the dev set as the corpus for the early stopping for efficiency.

For the QA reader, we initialize the model from the public TAPAS checkpoint.
We use a batch size of 512, train for 50,000 steps with a learning rate of 1e-6, and dropout rate 0.2.
In this setup we do not use early stopping but always train the model for the full number of steps.
We limit the maximal answer length to 10 word pieces.
The hyper-parameters of the QA model were optimized using a black box Bayesian optimizer similar to Google Vizier \cite{vizier}.
We used the hyper-parameter bounds given in Table \ref{tab:hp_bounds}.

\begin{table}[H]
\begin{center}
\begin{tabular}{ccc}
parameter & min & max \\
\toprule
learning rate & $1\mathrm{e}{-6}$ & $1\mathrm{e}{-2}$ \\
warm up ratio & $0.0$ & $0.2$ \\
dropout& $0.0$ & $0.2$\\
\bottomrule
\end{tabular}
\end{center}
\caption{Hyper-parameter ranges for tuning the QA model.}
\label{tab:hp_bounds}
\end{table}

We trained all models on 32 Cloud TPU v3.
Pre-Training a retrieval model takes approx. 6 days.
Training a retrieval model takes approx. 4-5h.
Training a QA model takes approx. 10h. 

The number of parameters is the same as for a BERT large model: 340M.

\section{Results}
\label{app:results}

Dev and test results for the retrieval experiments are given in Table \ref{tab:dev_test_retrieval}.
Dev and test results for end-to-end QA are given in the appendix in Table \ref{tab:dev_test_qa}.

\begin{table*}
\begin{center}
\resizebox{2\columnwidth}{!}{
\begin{tabular}{lrrrrrr}
      &   Dev &      &       & Test \\
Model &   R@1 &  R@10 & R@50 & R@1 & R@10 & R@50 \\
\toprule
BM25        & 17.13 & 42.13 & 57.21 & 16.77 & 40.06 & 58.39 \\
DTR-Text   & 28.68 \err{0.37} & 67.76 \err{0.56} & 85.75 \err{0.84} & 32.90 \err{0.57} & 72.00 \err{2.14} & 86.86 \err{1.25} \\
DTR-Schema & 29.38 \err{2.11} & 67.67 \err{0.28} & 85.29 \err{0.19} & 34.36 \err{1.09} & 74.24 \err{1.46} & 88.37 \err{0.47} \\
DTR-Text +hnbm25 & 35.30 \err{2.05} & 74.10 \err{1.50} & 87.85 \err{0.23} & 40.88 \err{0.66} & 78.10 \err{0.74} & 91.48 \err{0.21} \\
\midrule
DTR & 31.58 \err{0.09} & 71.79 \err{0.00} & 88.38 \err{0.37} & 36.24 \err{0.57} & 76.02 \err{0.10} & 90.25 \err{0.89} \\
DTR +hnbm25 &  37.86 \err{1.77} & 75.65 \err{0.72} & 89.58 \err{0.39} & 42.17 \err{2.91} & 80.51 \err{0.66} & 92.31 \err{0.70} \\
DTR +hn & 39.14 \err{0.98} & 76.13 \err{1.30} & 89.91 \err{0.35} & 42.42 \err{2.94} & 81.13 \err{1.61} & 92.56 \err{0.96} \\
\midrule
DTR -pt & 9.05 \err{1.08} & 35.73 \err{2.21} & 60.34 \err{3.20} & 16.64 \err{1.49} & 47.80 \err{1.42} & 68.68 \err{0.91} \\
\bottomrule
\end{tabular}
}
\end{center}
\caption{Table retrieval results on NQ-TABLES dev and test sets. hn: hard negatives, hnbm25: hard negatives from BM25 baseline, pt: pre-training.}
\label{tab:dev_test_retrieval}
\end{table*}

\begin{table*}
\begin{center}
\resizebox{2.0\columnwidth}{!}{
\begin{tabular}{llrrrrrrrr}
          &        & Dev&    &           &           & Test \\
Retriever & Reader & EM & F1 & Oracle EM & Oracle F1 & EM & F1 & Oracle EM & Oracle F1 \\
\toprule
BM25 & TAPAS & 18.76 \err{0.80} & 26.32 \err{1.23} & 25.57 \err{0.77} & 37.45 \err{0.81} & 21.46 \err{0.71} & 28.24 \err{0.78} & 29.51 \err{0.49} & 40.79 \err{0.43} \\
DTR-Text & Bert & 21.11 \err{1.09} & 29.17 \err{1.11} & 30.74 \err{1.07} & 42.82 \err{0.80} & 29.58 \err{0.85} & 37.38 \err{0.87} & 39.39 \err{0.31} & 51.48 \err{0.30} \\
DTR-Text & TAPAS & 27.67 \err{1.30} & 37.13 \err{1.74} & 36.44 \err{1.35} & 49.17 \err{1.66} & 33.78 \err{1.12} & 43.49 \err{1.23} & 42.83 \err{0.74} & 56.46 \err{0.55} \\
DTR-Text +hnbm25 & TAPAS & 27.84 \err{0.95} & 38.00 \err{1.14} & 38.97 \err{0.81} & 52.38 \err{0.97} & 36.89 \err{0.77} & 46.67 \err{0.98} & 46.30 \err{0.65} & 59.22 \err{0.84} \\
DTR-Schema & TAPAS & 27.12 \err{1.04} & 36.14 \err{1.17} & 36.19 \err{0.93} & 48.81 \err{1.29} & 32.75 \err{0.36} & 42.19 \err{0.21} & 42.63 \err{0.68} & 55.05 \err{0.59} \\
\midrule
DTR & TAPAS & 27.84 \err{1.62} & 37.77 \err{1.86} & 38.43 \err{0.75} & 51.26 \err{0.69} & 35.50 \err{0.45} & 45.44 \err{0.53} & 46.09 \err{0.47} & 59.01 \err{0.30} \\
DTR +hn & TAPAS & 28.67 \err{0.57} & 39.14 \err{0.46} & 39.38 \err{0.69} & 53.08 \err{0.43} & 37.69 \err{0.87} & 47.70 \err{1.05} & 48.20 \err{0.53} & 61.50 \err{0.34} \\
DTR +hn c=1 & TAPAS & 23.50 \err{0.12} & 33.44 \err{0.30} & 23.50 \err{0.12} & 33.44 \err{0.30} & 31.12 \err{0.31} & 39.44 \err{0.20} & 31.12 \err{0.31} & 39.44 \err{0.20} \\
DTR +hn c=50 & TAPAS & 23.97 \err{2.22} & 33.72 \err{2.81} & 42.11 \err{1.95} & 58.02 \err{2.08} & 30.73 \err{2.79} & 40.77 \err{3.26} & 47.26 \err{1.33} & 63.04 \err{1.51} \\
DTR +hnbm25 & TAPAS & 27.67 \err{0.63} & 37.77 \err{0.93} & 40.26 \err{0.94} & 53.77 \err{0.80} & 36.61 \err{0.76} & 46.74 \err{0.82} & 47.46 \err{0.83} & 60.72 \err{0.91} \\
\bottomrule
\end{tabular}
}
\end{center}
\caption{QA results on NQ-TABLES dev and test set. c: Number of candidates (default is 10), hn: With hard negatives, hnbm25: with hard negatives from BM25.}
\label{tab:dev_test_qa}
\end{table*}